\pdfoutput=1
\documentclass[11pt,oneside,english]{amsart}
\usepackage{mathpazo}
\usepackage[scaled=0.92]{helvet}
\usepackage{courier}
\usepackage[T1]{fontenc}
\usepackage[latin9]{inputenc}
\usepackage[a4paper]{geometry}
\usepackage{amstext}
\usepackage{amsthm}
\usepackage{amssymb}
\usepackage{graphicx}
\usepackage{setspace}
\usepackage[authoryear]{natbib}
\makeatletter

\providecommand{\tabularnewline}{\\}

\numberwithin{equation}{section}
\numberwithin{figure}{section}

\theoremstyle{plain}

\numberwithin{equation}{section}
\usepackage{tikz}
\usetikzlibrary{fit,positioning}

\makeatother

\usepackage{babel}
\begin{document}

\title[DOLDA supervised topic model]{DOLDA - a regularized supervised topic model for high-dimensional
multi-class regression}

\author{M{\aa}ns Magnusson, Leif Jonsson and Mattias Villani}

\thanks{Magnusson: \textit{Division of Statistics and Machine Learning, Dept.
of Computer and Information Science, SE-581 83 Linkoping, Sweden}.
\textit{E-mail: mans.magnusson@liu.se}. Jonsson: \textit{Ericsson
AB and Dept. of Computer and Information Science, SE-164 80 Stockholm,
Sweden}. \textit{E-mail:} leif.jonsson@ericsson.com. Villani: \textit{Division
of Statistics and Machine Learning, Dept. of Computer and Information
Science, SE-581 83 Linkoping, Sweden}. \textit{E-mail: mattias.villani@liu.se}.}
\begin{abstract}
Generating user interpretable multi-class predictions in data rich
environments with many classes and explanatory covariates is a daunting
task. We introduce Diagonal Orthant Latent Dirichlet Allocation (DOLDA),
a supervised topic model for multi-class classification that can handle
both many classes as well as many covariates. To handle many classes
we use the recently proposed Diagonal Orthant (DO) probit model \citep{Johndrow}
together with an efficient Horseshoe prior for variable selection/shrinkage
\citep{carvalho2010horseshoe}. We propose a computationally efficient
parallel Gibbs sampler for the new model. An important advantage of
DOLDA is that learned topics are directly connected to individual
classes without the need for a reference class. We evaluate the model's
predictive accuracy on two datasets and demonstrate DOLDA's advantage
in interpreting the generated predictions.
\end{abstract}

\maketitle

\section{Introduction}

During the last decades more and more textual data have become available,
creating a growing need to statistically analyze large amounts of
textual data. The hugely popular Latent Dirichlet Allocation (LDA)
model introduced by \citet{blei2003latent} is a generative probability
model where each document is summarized by a set of latent semantic
themes, often called topics; formally, a topic is a probability distribution
over the vocabulary. An estimated LDA model is therefore a compressed
latent representation of the documents. LDA is a mixed membership
model where each document is a mixture of topics, where each word
(token) in a document belongs to a single topic. The basic LDA model
is unsupervised, i.e. the topics are learned solely from the words
in the documents without access to document labels.

In many situations there are also other information we would like
to incorporate in modeling a corpus of documents. A common example
is when we have labeled documents, such as ratings of movies together
with a movie description, illness category in medical journals or
the location of the identified bug together with bug reports. In these
situation, one can use a so called supervised topic model to find
the semantic structure in the documents that are related to the class
of interest. One of the first approaches to supervised topic models
was proposed by \citet{mcauliffe2008supervised}. The authors propose
a supervised topic model based on the generalized linear model framework,
thereby making it possible to link binary, count and continuous response
variables to topics that are inferred jointly with the regression/classification
effects. In this approach the semantic content of a text in the form
of topics predicts the response variable $y$. This approach is often
referred to as \emph{downstream} supervised topic models, contrary
to an \emph{upstream} supervised approach where the label $y$ governs
how the topics are formed, see e.g. \citet{ramage2009labeled}. 

Many downstream supervised topic models have been studied, mainly
in the machine learning literature. \citet{mcauliffe2008supervised}
focus on downstream supervision using generalized linear regression
models. \citet{jiang2012monte} propose a supervised topic model using
a max-margin approach to classification and \citet{zhu2013improved}
propose a logistic supervised topic model using data augmentation
with polya-gamma variates . \citet{perotte2011hierarchically} use
a hierarchical binary probit model to model a hierarchical label structure
in the form of a binary tree structure. 

Most of the proposed supervised topic models have been motivated by
trying to find good classification models and the focus has naturally
been on the predictive performance. However, the predictive performance
of most supervised topic models are just slightly better than using
a Support Vector Machine (SVM) with covariates based on word frequencies
\citep{jameel2015supervised}. While predictive performance is certainly
important, the real attraction of supervised topic models comes from
their ability to learn semantically relevant topics and to use those
topics to produce accurate \emph{interpretable} predictions of documents
or textual data. The interpretability of a model is an often neglected
feature, but is crucial in real world applications. As an example,
\citet{parnin2011automated} show that bug fault localization systems
are quickly disregarded when the users can not understand how the
system has reached the predictive conclusion. Compared to other text
classification systems, topic models are very well suited for interpretable
predictions since topics are an abstract entity that are possible
for humans to grasp. The problems of interpretability in multi-class
supervised topic models can be divided into three main areas. 

First, most supervised topic models use a logit or probit approach
where the model is identified by the use of a reference category,
to which the effect of any covariate is compared. This defeats one
of the main purposes of supervised topic models since this complicates
the interpretability of the models.

Second, to handle multi-class categorization a topic should be able
to affect multiple classes, and some topics may not influence any
class at all. In most supervised topic modeling approaches (such as
\citealt{jiang2012monte,zhu2013improved,jameel2015supervised}) the
multi-class problem is solved using binary classifiers in a ``one-vs-all''
classification approach. This approach works well in the situation
of evenly distributed classes, but may not work well for skewed class
distributions \citet{rubin2012statistical}. A one-vs-all approach
also makes it more difficult to interpret the model. Estimating one
model per class makes it impossible to see which classes that are
affected by the same topic and which topics that do not predict any
label. In these situations we would like to have \textit{one} topic
model to interpret. The approach of one-vs-all predictions are also
costly from an estimation point of view since we need to estimate
one model per class \citet{Zheng:2015:LTS:2783258.2783371}. 

Third, there can be situations with hundreds of classes and hundreds
of topics (see \citet{Jonsson2016} for an example). Without regularization
or variable selection we would end up with a model with too many parameters
to interpret and very uncertain parameter estimates. In a good predictive
supervised topic model one would like to find a small set of topics
that are strong determinants of a single document class label. This
is especially relevant when the number of observations in different
classes are skewed, a problem common in real world situations \citep{rubin2012statistical}.
In the more rare classes we would like to induce more shrinkage while
in the situation with more data we would like to have less shrinkage
in the model.

Multi-class regression is a non-trivial problem in Bayesian modeling.
Historically, the multinomial probit model has been preferred due
to the data augmentation approach proposed by \citet{albert1993bayesian}.
Augmenting the sampler using latent variables lead to straight forward
Gibbs sampling with conditionally-conjugate updates of the regression
coefficients. The Albert-Chib sampler often tend to mix slowly, and
the same holds for improved sampler such as the parameter expansion
approach in \citet{imai2005bayesian}. Recently, a similar data augmentation
approach using polya-gamma variables is proposed for the Bayesian
logistic regression model by \citet{polson2013bayesian}. This approach
preserve the conditional-conjugacy in the case of a Normal prior for
the regression coefficients and has been the foundation for the supervised
topic model in \citet{zhu2013improved}.

In this paper we explore a new approach to supervised topic models
that produce accurate multi-class predictions from semantically interpretable
topics using a fully Bayesian approach, hence solving all three of
the above mentioned problems. The model combines LDA with the recently
proposed Diagonal Orthant (DO) probit model \citet{Johndrow} for
multi-class classification with an efficient Horseshoe prior that
achieves sparsity and interpretation by aggressive shrinkage \citep{carvalho2010horseshoe}.
The new Diagonal Orthant Latent Dirichlet Allocation (DOLDA)\footnote{DOLDA is Swedish for hidden or latent.}
model is demonstrated to have competitive predictive performance yet
producing interpretable multi-class predictions from semantically
relevant topics.  

\section{Diagonal Orthant Latent Dirichlet Allocation}

\subsection{Handling the challenges for high-dimensional interpretable supervised
topic models}

To solve the first and second challenge identified in the Introduction,
reference classes and multi-class models, we propose to use the Diagonal
Orthant (DO) probit model in \citet{Johndrow} as an alternative to
the multinomial probit and logit models. \citet{Johndrow} propose
a Gibbs sampler for the model and shows that it mixes well. One of
the benefits of the DO model is that all classes can be independently
modeled using binary probit models when conditioning on the latent
variable, thereby removing the need for a reference class. The parameters
of the model can be interpreted as the effect of the covariate on
the marginal probability of a specific class, which make this model
especially attractive when it comes to interpreting the inferred topics.
This model also include multiple classes in an efficient way that
makes it possible to estimate a multi-class linear model in parallel
over the classes. 

The third problem of modeling supervised topic models is that the
semantic meanings of all topics do not necessarily have an effect
on our label of interest; one topic may have an effect on one or more
classes, and some topics may just be noise that we do not want to
use in the supervision. In the situation with many topics and many
classes we will also have a very large number of parameters to analyze.
The Horseshoe prior in \citet{carvalho2010horseshoe} was specifically
designed to filter out signals from massive noise. This prior uses
a local-global shrinkage approach to shrink some (or most) coefficients
to zero while allowing for sparse signals to be estimated without
any shrinkage. This approach has shown good performance in linear
regression type situations \citep{castillo2015}, something that makes
it straight forward to incorporate other covariates into our model,
which is rarely done in the area of supervised topic models. Different
global shrinkage parameters are used for the different classes to
handle the problem with unbalanced number of observations in different
classes. This makes it possible to shrink more when there are less
data for a given class and shrink less in classes with more observations.

\subsection{Generative model\label{subsec:Generative-model}}

\begin{table}
\begin{centering}
\begin{tabular}{clccl}
\hline 
{\scriptsize{}Symbol} & {\tiny{}Description} &  & {\tiny{}Symbol} & {\tiny{}Description}\tabularnewline
\cline{1-2} \cline{4-5} 
{\scriptsize{}$\mathcal{V}$} & {\tiny{}The set of word types/vocabulary} &  & {\tiny{}$\beta$} & {\tiny{}The prior for $\Phi$: $K\times V$}\tabularnewline
{\scriptsize{}$V$} & {\tiny{}The size of the vocabulary i.e $V=|\mathcal{V}|$} &  & {\tiny{}$\Theta$} & {\tiny{}Document-topic proportions: $D\times K$}\tabularnewline
{\scriptsize{}$v$} & {\tiny{}Word type} &  & {\tiny{}$\theta_{d}$} & {\tiny{}Topic probability for document $d$}\tabularnewline
{\scriptsize{}$\mathcal{K}$} & {\tiny{}The set of topics} &  & {\tiny{}$\alpha$} & {\tiny{}The prior for $\Theta$: $D\times K$}\tabularnewline
{\scriptsize{}$K$} & {\tiny{}The number of topics i.e $K=|\mathcal{K}|$} &  & \textbf{\tiny{}$\mathbf{M}$} & {\tiny{}\#of topics indicators in each document by topics: $D\times K$}\tabularnewline
{\scriptsize{}$L$} & {\tiny{}The number of labels/categories} &  & \textbf{\tiny{}a} & {\tiny{}Matrix of latent gaussian variables: $D\times L$ }\tabularnewline
{\scriptsize{}$\mathcal{L}$} & {\tiny{}The set of labels/categories} &  & {\tiny{}$\eta$} & {\tiny{}Coefficient matrix for each label and covariate: $(K+P)\times L$}\tabularnewline
{\scriptsize{}$D$} & {\tiny{}\#of observations/documents i.e. $D=|\mathcal{D}|$} &  & {\tiny{}$\eta_{0}$ } & {\tiny{}Prior for $\eta$: $L\times(K+P)$}\tabularnewline
{\scriptsize{}$\mathcal{D}$} & {\tiny{}The set of observations/documents} &  & {\tiny{}$z_{n,d}$} & {\tiny{}Topic indicator for token $n$ in document $d$}\tabularnewline
{\scriptsize{}$P$} & {\tiny{}The number of non-textual covariates/features} &  & {\tiny{}$\bar{\mathbf{z}}$} & {\tiny{}Proportion of topic indicators by document: $D\times K$}\tabularnewline
{\scriptsize{}$N$} & {\tiny{}The total number of tokens} &  & {\tiny{}$w_{n,d}$} & {\tiny{}Token $n$ in document $d$}\tabularnewline
{\scriptsize{}$N_{d}$} & {\tiny{}The number of tokens in document $d$} &  & {\tiny{}$\mathbf{w}_{d}$} & {\tiny{}Vector of tokens in document $d$: $1\times N_{d}$}\tabularnewline
{\scriptsize{}$\mathbf{N}$} & {\tiny{}\# obs topic-word type indicators: $K\times V$} &  & {\tiny{}$y_{d}$} & {\tiny{}Label for document $d$}\tabularnewline
{\scriptsize{}$\Phi$} & {\tiny{}The matrix with word-topic probabilities : $K\times V$} &  & {\tiny{}$\mathbf{X}$} & {\tiny{}Covariate/feature matrix (including intercept): $D\times P$}\tabularnewline
{\scriptsize{}$\phi_{k}$} & {\tiny{}The word probabilities for topic $k$: $1\times V$} &  & {\tiny{}$\mathbf{x}_{d}$} & {\tiny{}Covariate/features for document $d$}\tabularnewline
\hline 
\end{tabular}
\par\end{centering}
\caption{DOLDA model notation. \label{tab:Notation-used.}}
\end{table}

\begin{figure}
\begin{tikzpicture} 
\tikzstyle{main}=[circle, minimum size = 10mm, thick, draw =black!80, node distance = 16mm] 
\tikzstyle{connect}=[-latex, thick] 
\tikzstyle{box}=[rectangle, draw=black!100]
\node[main, draw=none, fill=none] (alpha) {$\alpha$};
\node[main] (theta) [right=of alpha,xshift=-8mm,label=below:$\theta$] { };
\node[main] (z) [right=of theta,label=above right:$\mathbf{z}$] {};
\node[main, fill = black!10] (w) [right=of z,label=below:$\mathbf{w}$] { };
\node[main] (phi) [right=of w,label=below:$\Phi$] { };
\node[main, fill = black!10] (x) [below=of z, yshift=8mm, label=below:$\mathbf{x}$] { };
\node[main, fill = black!10] (y) [below=of x, yshift=8mm,label=below:$y$] { };
\node[main] (a) [right=of y,label=above right:$\mathbf{a}$] { };
\node[draw=none,fill=none,right=of phi] (gamma) {$\gamma$};
\node[main] (eta) [right=of a,label=right:$\eta$] { };
\node[main, above=of eta,xshift=3mm, yshift=-13mm,label=above:$\lambda$] (lambda) {};
\node[main, above=of eta,xshift=-9mm, yshift=-13mm,label=above:$\tau$] (tau) {};
\path 
(alpha) edge [connect] (theta)
(theta) edge [connect] (z)         
(z) edge [connect] (w)
(x) edge [connect] (a)
(z) edge [connect] (a)
(eta) edge [connect] (a)
(a) edge [connect] (y)
(gamma) edge [connect] (phi)           
(tau) edge [connect] (eta) 
(lambda) edge [connect] (eta) 
(phi) edge [connect] (w);
\node[rectangle, inner sep=0mm, fit= (z) (w),label=below right:$N$, yshift=1mm, xshift=11.5mm] {};
\node[rectangle, inner sep=-1mm, fit= (phi),label=below right:$K$, xshift=0mm] {};
\node[rectangle, inner sep=0mm, xshift=-7mm, fit= (theta) (z) (w) (x) (y), label=below left:$D$] {};
\node[rectangle, inner sep=5mm, xshift=-1mm, yshift=7mm,fit= (a) (eta) (tau) (lambda) ,label=below right:$L$] {};
\node[rectangle, inner sep=4.4mm,draw=black!100, fit= (z) (w)] {};
\node[rectangle, inner sep=5.5mm, draw=black!100, fit = (theta) (z) (w) (x) (y)] {};
\node[rectangle, inner sep=5mm, draw=black!100, fit = (phi)] {};
\node[rectangle, inner sep=5mm, xshift=-3mm, yshift=1mm, draw=black!100, fit = (a) (eta) (tau) (lambda)] {};
\end{tikzpicture}

\caption{The Diagonal Orthant probit supervised topic model (DOLDA)}
\label{Fig:GraphicalDOLDA}
\end{figure}
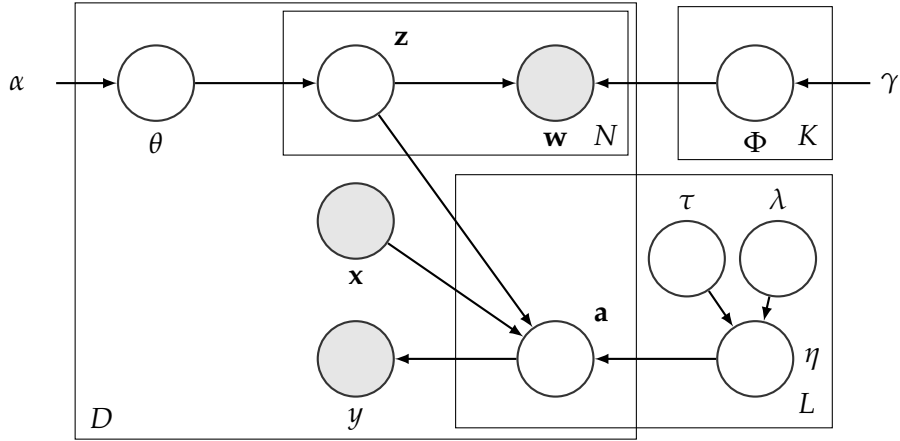

The generative model is described below. See also a graphical description
of the model in Figure \ref{Fig:GraphicalDOLDA}. A summary of the
notation is given in Table \ref{tab:Notation-used.}
\begin{enumerate}
\item For each topic $k=1,...,K$

\begin{enumerate}
\item Draw a distribution over words $\phi_{k}\sim\mbox{Dir}_{V}(\gamma_{k})$
\end{enumerate}
\item For each label $l\in L$

\begin{enumerate}
\item Draw a global shrinkage parameter $\tau_{l}\sim C^{+}(0,1)$
\item Draw local shrinkage parameters for the $p$th covariate $\lambda_{l,p}\sim C^{+}(0,1)$
\item Draw coefficients\footnote{The intercept is estimated using a normal prior.}
$\eta_{l,p}\sim\mathcal{N}_{K+P}(0,\tau_{l}^{2}\lambda_{l,p}^{2})$
\end{enumerate}
\item For each observation/document $d=1,...,D$

\begin{enumerate}
\item Draw topic proportions $\theta_{d}|\alpha\sim\mbox{Dir}_{K}(\alpha)$
\item For $n=1,...,N_{d}$ 

\begin{enumerate}
\item Draw topic assignment $z_{n,d}|\theta_{d}\sim\mbox{Categorical}(\theta_{d})$ 
\item Draw word $w_{n,d}|z_{n,d},\phi_{z_{n,d}}\sim\mbox{Categorical}(\phi_{z_{n,d}})$
\end{enumerate}
\item $y_{d}\sim\mbox{Categorical}(\mathbf{p}_{d})$ where
\[
\mathbf{p}_{d}=\left[\sum_{l}^{L}F_{l}^{\mathcal{N}(0,1)}\left((\bar{\mathbf{z}},\mathbf{x})_{d}^{\top}\eta_{l\cdot}\right)\right]^{-1}\left(F_{1}^{\mathcal{N}(0,1)}\left((\bar{\mathbf{z}},\mathbf{x})_{d}^{\top}\eta_{1\cdot}\right),...,F_{L}^{\mathcal{N}(0,1)}\left((\bar{\mathbf{z}},\mathbf{x})_{d}^{\top}\eta_{L\cdot}\right)\right)
\]
and $F_{l}()$ is the univariate normal CDF \citep{Johndrow}.
\end{enumerate}
\end{enumerate}

\section{Inference}

\subsection{The MCMC algorithm}

Markov Chain Monte Carlo (MCMC) is used to estimate the model parameters.
We use different global shrinkage parameters $\tau_{l}$ for each
class, motivated by the fact that the different classes can have different
number of observations. This gives the following sampler for inference
in DOLDA.
\begin{enumerate}
\item Sample the latent variables $a_{d,l}^{(i)}\sim\mbox{\ensuremath{\mathcal{N}}}_{+}((\mathbf{x}\mbox{ }\bar{\mathbf{z}})_{d}^{T}\eta_{l},1)$
for $l=y_{d}$ and $a_{d,l}\sim\mbox{\ensuremath{\mathcal{N}}}_{-}((\mathbf{x}\mbox{ }\bar{\mathbf{z}})_{d}^{T}\eta_{l},1)$
for $l\neq y_{d}$, where $\mathcal{N}_{+}$ and $\mathcal{N}_{-}$
is the positive and negative truncated normal distribution, truncated
at 0.
\item Sample all the regression coefficients as in an ordinary Bayesian
linear regression per class label $l$ where $\eta_{l}\sim\mbox{\ensuremath{\mathcal{MVN}}}\left(\mu_{l},((\mathbf{X}\mbox{ }\bar{\mathbf{z}})^{T}\mathbf{(\mathbf{X}\mbox{ }\bar{\mathbf{z}})}+\tau_{l}^{2}\Lambda_{l})^{-1}\right)$
and $\Lambda_{l}$ is a diagonal matrix with the local shrinkage parameters
$\lambda_{i}$ per parameter in $\eta_{l}$ and $\mu_{l}=((\mathbf{X}\mbox{ }\bar{\mathbf{z}})^{T}(\mathbf{X}\mbox{ }\bar{\mathbf{z}})+\tau_{l}^{2}\Lambda_{l})^{-1}(\mathbf{X}\mbox{ }\bar{\mathbf{z}})^{T}\mathbf{a}_{l}$
\item Sample the global shrinkage parameters $\tau_{l}$ at iteration $j$
using the following two step slice sampling:
\begin{eqnarray*}
u & \sim & \mathcal{U}\left(0,\left[1+\frac{1}{\tau_{l,(j-1)}}\right]{}^{-1}\right)\\
\frac{1}{\tau_{l,j}^{2}} & \sim & \mbox{\ensuremath{\mathcal{G}}}\left((p+1)/2,\frac{1}{2}\sum_{p=1}^{P}\left(\frac{\eta_{l,p}}{\lambda_{l,p}}\right)^{2}\right)I\left[\frac{1}{\tau_{l,(j-1)}^{2}}<(1-u)/u\right]
\end{eqnarray*}
where $I$ indicates the truncation region for the truncated gamma.
\item Sample each local shrinkage parameter $\lambda_{i,l}$ as 
\begin{eqnarray*}
u & \sim & \mathcal{U}\left(0,\left[1+\frac{1}{\lambda_{p,l,(j-1)}^{2}}\right]{}^{-1}\right)\\
\frac{1}{\lambda_{p,l,j}^{2}} & \sim & \mathcal{E}\left(\frac{1}{2}\left(\frac{\eta_{l,p}}{\tau_{l}}\right)^{2}\right)I\left[\frac{1}{\lambda_{p,l,(j-1)}^{2}}<(1-u)/u\right]
\end{eqnarray*}
\item Sample the topic indicators $\mathbf{z}$ 
\begin{eqnarray*}
p(z_{i,d}=k|w_{i},\mathbf{z}^{\lnot i},\eta,\mathbf{a}) & \propto & \phi_{v,k}\cdot\left(n_{d,k}^{(d),\lnot i}+\alpha\right)\cdot\\
 &  & \exp\left(-\frac{1}{2}\sum_{l}^{L}\left[-2\frac{\eta_{l,k}}{N_{d}}\left(a_{d,l}-(\bar{\mathbf{z}}_{d}^{\lnot i}\mbox{ }\mathbf{x}_{d})\eta_{l}^{\intercal}\right)+\left(\frac{\eta_{l,k}}{N_{d}}\right)^{2}\right]\right)
\end{eqnarray*}
where $n^{(d)}$ is a $D\times K$ count matrix containing the sufficient
statistics for $\Theta$.
\item Sample the topic-vocabulary distributions $\Phi$
\[
\phi_{k}\sim\mbox{Dir}(\beta_{k}+n_{k}^{(w)})
\]
where $n^{(w)}$ is a $K\times V$ count matrix containing the sufficient
statistics for $\Phi$. 
\end{enumerate}

\subsection{Efficient parallel sampling of $\mathbf{z}$}

To improve the speed of the sampler we cache the calculations done
in the supervised part of the topic indicator sampler and parallelize
the sampler. Some commonly used text corpora have several hundreds
of millions topic indicators, so efficient sampling of the $\mathbf{z}$
are absolutely crucial in practical applications. The basic sampler
for $\mathbf{z}$ can be slow due to the serial nature of the collapsed
sampler and the fact that the supervised part of $p(z_{i,d})$ involves
a dot product. 

The supervised part of document $d$ can be expressed as $\mbox{\ensuremath{\exp}}\left(g_{d,k}^{\lnot i}\right)$
where 
\[
g_{d,k}^{\lnot i}=-\frac{1}{2}\sum_{l}^{L}\left[-2\frac{\eta_{l,k}}{N_{d}}\left(a_{d,l}-(\bar{\mathbf{z}}_{d}^{\lnot i}\mbox{ }\mathbf{x}_{d})\eta_{l}^{\intercal}\right)+\left(\frac{\eta_{l,k}}{N_{d}}\right)^{2}\right].
\]
By realizing that sampling a topic indicator just means updating a
small part of this equation we can derive the relationship 
\[
g_{d,k}=g_{d,k}^{\lnot i}-\frac{1}{N_{d}^{2}}\sum_{l}^{L}\eta_{l,k}\eta_{l,z_{i}}
\]
where the expression $\sum_{l}^{L}\eta_{l,k}\eta_{l,z_{i}}$ can be
calculated once per iteration in $\eta$ and be stored in a two-dimensional
array of size $K^{2}$. We can then use the above relationship to
update the supervision after sampling each topic indicator by calculating
$g_{d,k}^{\lnot i}$ ``on the fly'' based on the previous supervised
contribution $g_{d,k}^{\lnot(i-1)}$ in the following way 
\[
g_{d,k}^{\lnot i}=g_{d,k}^{\lnot(i-1)}+\frac{1}{N_{d}^{2}}\left[\sum_{l}^{L}\eta_{l,k}\eta_{l,z_{i}}-\sum_{l}^{L}\eta_{l,k}\eta_{l,z_{(i-1)}}\right]
\]
Caching $g_{d,k}^{\lnot i}$ leads to an order of magnitude speed
up for a model with 100 topics. 

To further improve the performance we parallelize the sampler and
use that documents are conditionally independent given $\Phi$. By
sampling $\Phi$, instead of marginalizing it out, we reduce the efficiency
of the MCMC somewhat, but we will converge to the true posterior and
the gain from parallelization is usually far greater than the reduced
efficiency \citep{magnusson2015parallelizing}. 

In summary, we have the following sampler for $z_{i,d}$
\begin{eqnarray*}
p(z_{i,d}=k|\cdot) & \propto & \phi_{k,v}\cdot\left(n_{d,k}^{(d),\lnot i}+\alpha\right)\cdot\exp\left(g_{d,k}^{\lnot i}\right).
\end{eqnarray*}
that can be sampled in parallel over the documents, and the elements
in $\Phi$ can be sampled in parallel over topics. The code is publicly
available at \textbf{https://github.com/lejon/DiagonalOrthantLDA}.

It is also straightforward to use the recently proposed cyclical Metropolis-Hastings
proposals in \citet{Zheng:2015:LTS:2783258.2783371} for inference
in DOLDA. The additional sampling of $\lambda_{p,l}$ and $\tau_{l}$
in our model can be done in $O(K+P)$ and is hence not affecting the
overall complexity of the sampler. But, as shown in \citet{magnusson2015parallelizing},
it is not obvious that the reduction in sampling complexity will result
in a faster sampling when MCMC efficiency is taken into account.

\subsection{Evaluation of convergence and prediction}

We evaluate the convergence of the MCMC algorithm by monitoring the
log-likelihood over the iterations:

\begin{eqnarray*}
\log\mathcal{L}(\mathbf{w},\mathbf{y}) & = & \sum_{d}^{D}\log\left[\sum_{s=1}^{J}(1-\mbox{cdf}_{\mathcal{N}}(-(\bar{\mathbf{z}}_{d}\mbox{ }\mathbf{x}_{d})\eta_{j}^{\intercal}))\prod_{l\neq s}\mbox{cdf}_{\mathcal{N}}(-(\bar{\mathbf{z}}_{d}\mbox{ }\mathbf{x}_{d})\eta_{l}^{\intercal})\right]\\
 &  & +\underbrace{\log p(\mathbf{w})}_{\mbox{LDA marginal LL}}
\end{eqnarray*}

To make predictions for a new document we first need to sample the
topic indicators of the given document from
\[
p(z_{i}=k|\mathbf{w}_{new},\Phi)\text{\ensuremath{\propto}}\bar{\phi}_{k,v}\cdot\left(\mathbf{M}_{d,k}+\alpha\right),
\]
where $\bar{\phi}_{k,v}$ is the mean of the last part of the posterior
draws of $\Phi$. We use the posterior mean based on the last iterations
instead of integrating out $\Phi$ to avoid potential problems with
label switching. However, we have not seen any indications of label
switching after convergence in our experiment, probably because the
data sets used for document predictions are usually quite large. The
topic indicators are sampled for the predicted document using the
fast PC-LDA sampler in \citet{magnusson2015parallelizing}. The mean
of the sampled topic indicator vector for the predicted document,
$\bar{\mathbf{z}}_{\text{new}}$, is then used for class predictions:

\[
y_{\mbox{pred}}=\arg\max\left((\bar{\mathbf{z}}_{\mbox{new}},\mathbf{x}_{\mbox{new}})^{\top}\eta\right).
\]
This is a maximum a posteriori estimate, but it is straightforward
to calculate the whole predictive distribution for the label.

\section{Experiments}

We collected a dataset containing the 8648 highest rated movies at
IMDb.com. We use both the textual description as well as information
about producers and directors to classify a given movie to a genre.
We also analyze the classical 20 Newsgroup dataset to compare the
accuracy with state-of-the-art supervised models. Our companion paper
\citep{Jonsson2016} applies the DOLDA model developed here to bug
localization in a large scale software engineering context using a
dataset with 15 000 bug reports each belonging to one of 118 classes.
We evaluate the proposed topic model with regard to accuracy and distribution
of regression coefficients. The experiments are performed on 2 sockets
with 8-core Intel Xeon E5-2660 Sandy Bridge processors at 2.2GHz and
32 GB DDR3 1600 memory at the National Super Computer centre (NSC)
at Link�ping University. 

\subsection{Data and priors}

The datasets are tokenized and a standard stop list of English words
are removed, as well as the most rare word types that makes up of
1 \% of the total tokens; we only include genres with at least 10
movies. 

\begin{table}
\begin{centering}
\begin{tabular}{|c|c|c|c|c|}
\hline 
Dataset & Classes ($L$) & Vocabulary ($V$) & Documents ($D$) & Tokens ($N$)\tabularnewline
\hline 
\hline 
IMDb & 20 & 7 530 & 8 648 & 307 569\tabularnewline
\hline 
20 Newsgroups & 20 & 23 941 & 15 077 & 2 008 897\tabularnewline
\hline 
\end{tabular}
\par\end{centering}
\caption{Datasets used in experiment}
\end{table}

In all experiments we used a relative vague priors setting $\alpha=\beta=0.01$
for the LDA part of the model and $c=100$ for the prior variance
of the $\eta$ coefficients in the normal model and for the intercept
coefficient when using the Horseshoe prior. The accuracy experiment
for IMDb was conducted using $5$-fold cross validation and the 20
Newsgroups corpus used the same train and test set as in \citet{zhu2012medlda}
to enable direct comparisons of accuracy. In the analysis of the IMDb
dataset no cross-validation was conducted, instead the whole data
set was used for estimation.

\subsection{Results}

\subsubsection*{20 Newsgroups}

\begin{figure}
\includegraphics{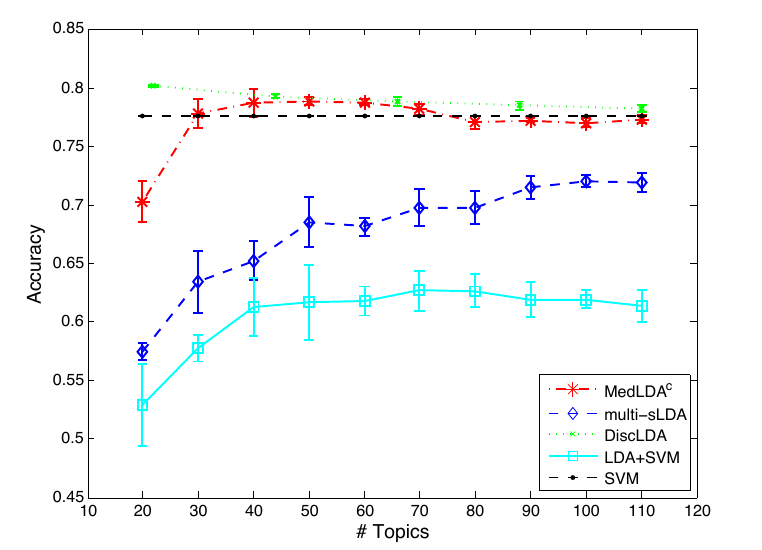}\includegraphics[scale=0.1]{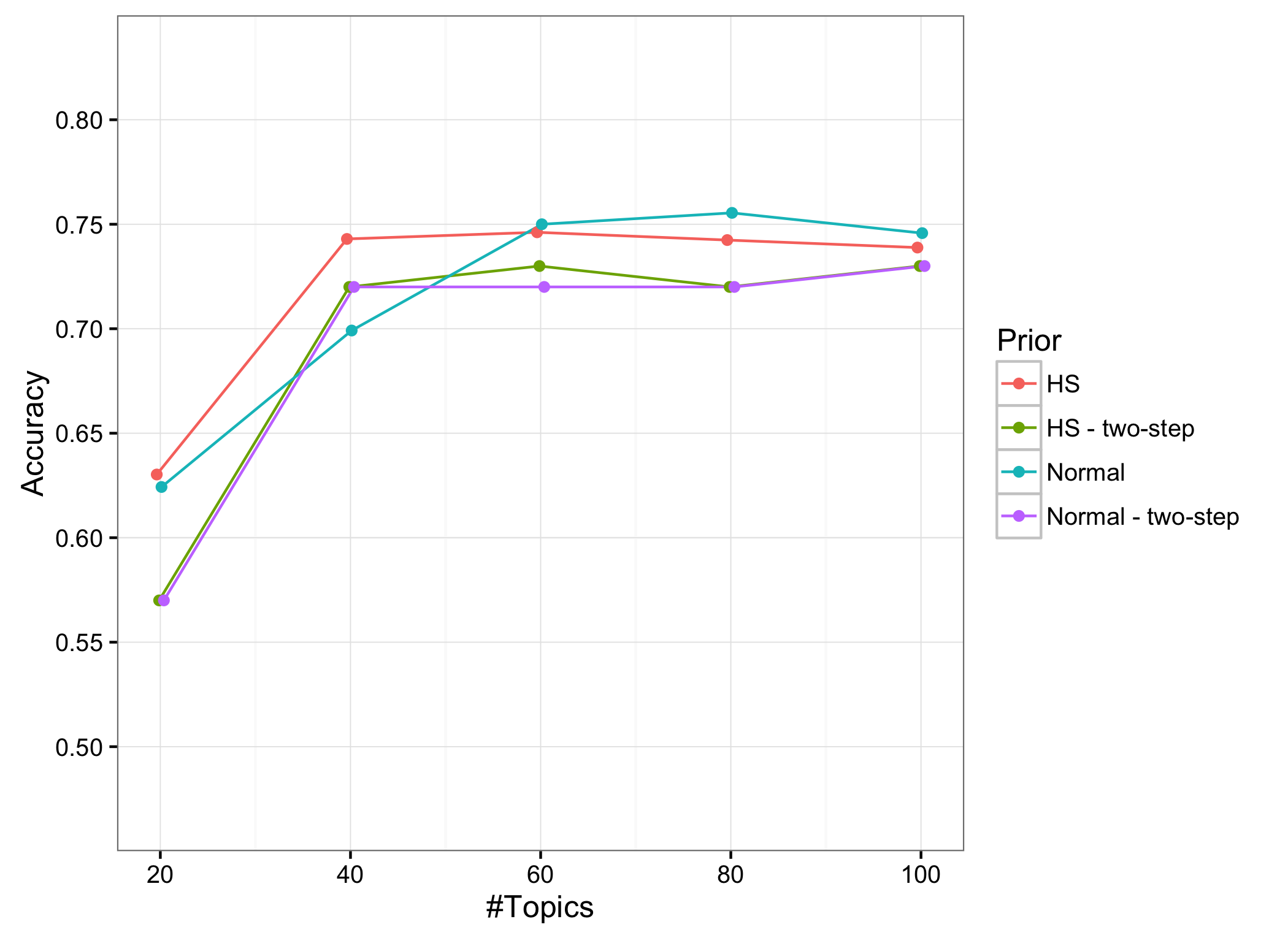}

\caption{Accuracy of MedLDA, taken from \citealt{zhu2012medlda} (left) and
accuracy of DOLDA for the 20 Newsgroup test set (right). \label{fig:Accuracy-of-MedLDA}}
\end{figure}

Figure \ref{fig:Accuracy-of-MedLDA} displays the accuracy on the
hold-out test set for the 20 Newsgroups dataset for different number
of topics. The accuracy of our model is slightly lower than MedLDA
and SVM using only textual features, but higher than both the classical
supervised multi-class LDA and the ordinary LDA together with an SVM
approach.

We can also see from Figure \ref{fig:Accuracy-of-MedLDA} that the
accuracy of using the DOLDA model with the topics jointly estimated
with the supervision part outperforms a two-step approach of first
estimating LDA and then using the DO probit model with the pre-estimated
mean topic indicators as covariates. This is true for both the Horseshoe
prior and the normal prior, but the difference is just a few percent
in accuracy.

The advantage of DOLDA is that it produces interpretable predictions
with semantically relevant topics. It is therefore reassuring that
DOLDA can compete in accuracy with other less interpretable models,
even when the model is dramatically simplified by aggressive Horseshoe
shrinkage for interpretational purposes. Our next data set illustrates
the interpretational strength of DOLDA. See also our companion paper
\citep{Jonsson2016} in the software engineering literature for further
demonstrations of DOLDAs ability to produce interpretable predictions
in industrial applications without sacrificing prediction accuracy.

\subsubsection*{IMDb}

\begin{figure}
\begin{centering}
\includegraphics[scale=0.1]{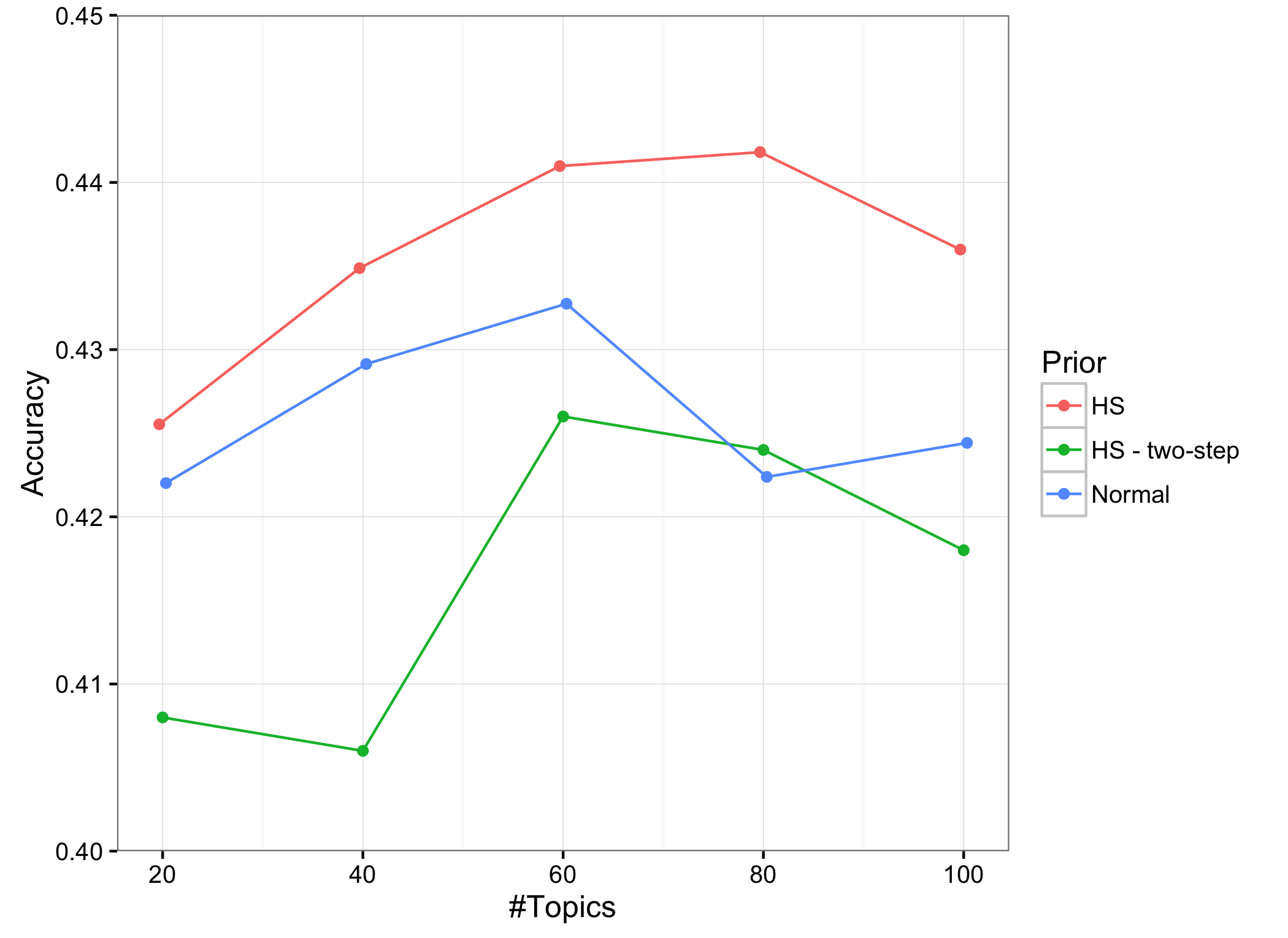}
\par\end{centering}
\caption{Accuracy for DOLDA on the IMDb data with normal and Horseshoe prior
and using a two step approach with the Horseshoe prior. \label{fig:Accuracy-for-DO}}
\end{figure}

Figure \ref{fig:Accuracy-for-DO} displays the accuracy on the IMDb
dataset as a function of the number of topics. The estimated DOLDA
model also contains several other discrete covariates, such as the
film's director and producer. The accuracy of the more aggressive
Horseshoe prior is better than the normal prior for all topic sizes.
A supervised approach with topics and supervision inferred jointly
is again outperforming a two-step approach. 

The Horseshoe prior gives somewhat higher accuracy than the normal
prior, and incorporating the Horseshoe prior let us handle many additional
covariates since the shrinkage prior will act as a type of variable
selection.

\begin{figure}
\begin{centering}
\includegraphics[scale=0.08]{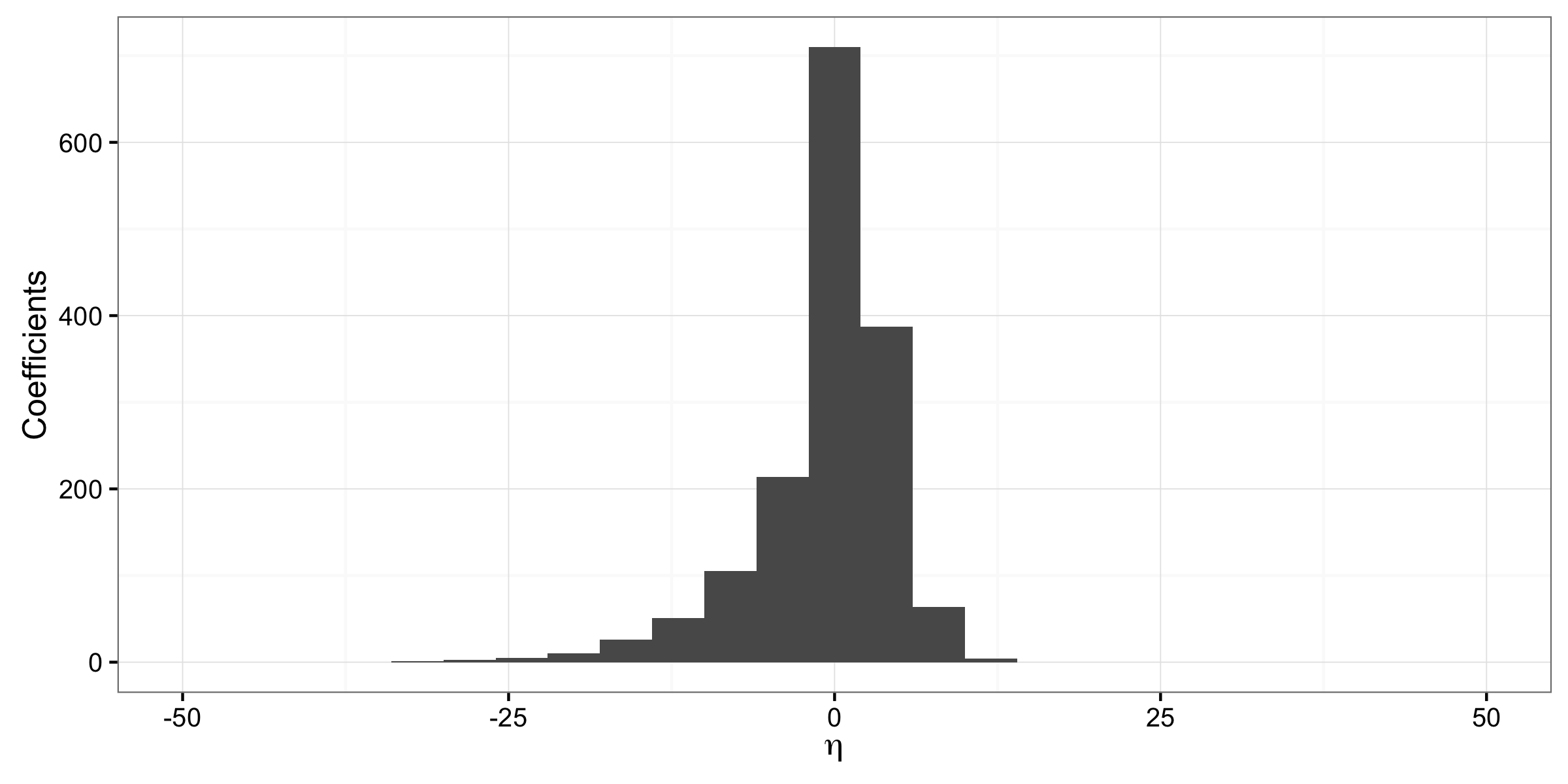}\includegraphics[scale=0.08]{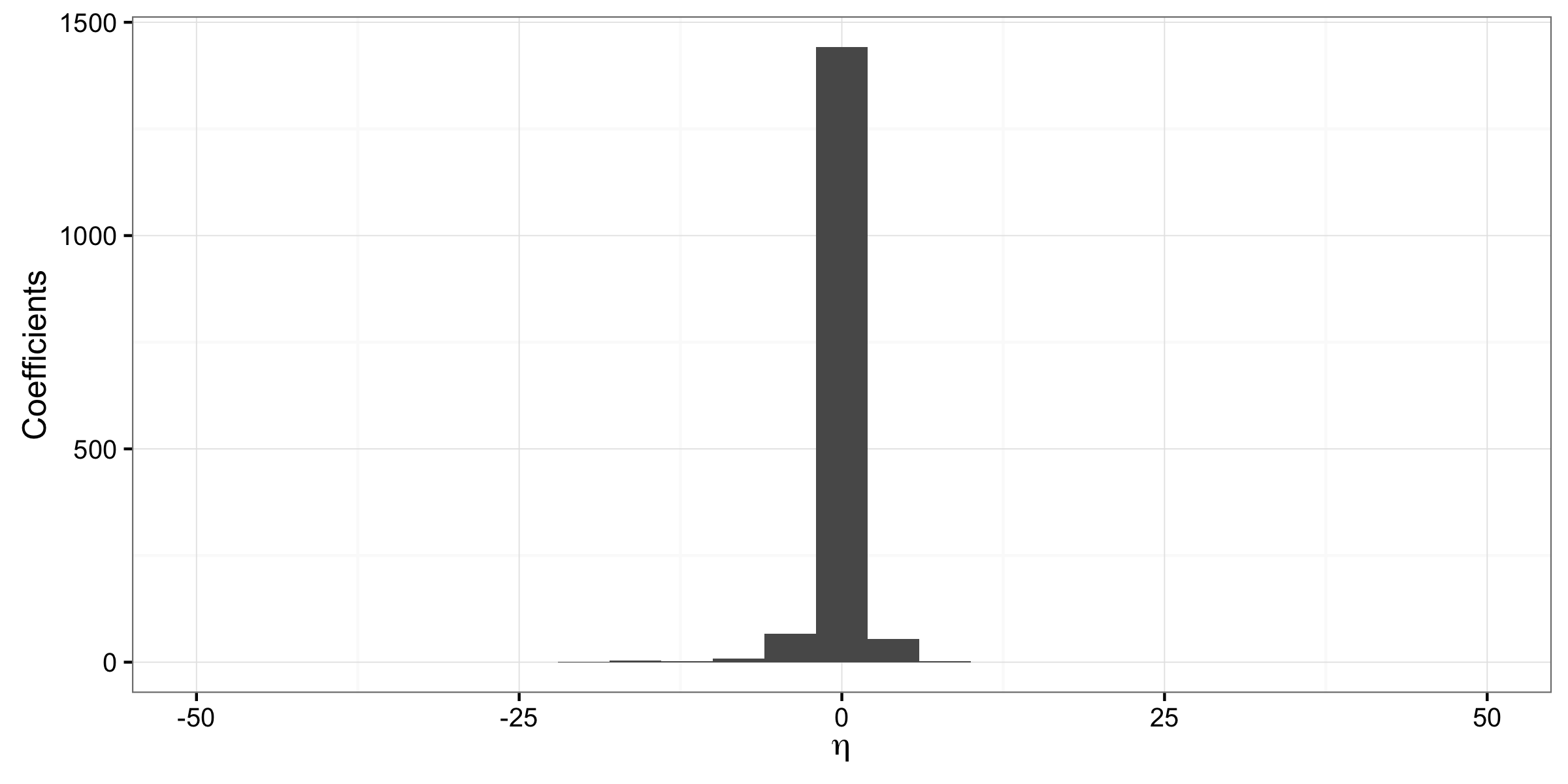}
\par\end{centering}
\caption{Coefficients for the IMDb dataset with 80 topics using the normal
prior (left) and the Horseshoe prior (right). \label{fig:Hist-beta}}
\end{figure}

To illustrate the interpretation of DOLDA we fit a new model using
only topics as covariates. Note first in Figure \ref{fig:Hist-beta}
how the Horseshoe prior is able to distinguish between so called signal
topics and noise topics; the Horseshoe prior is aggressively shrinking
a large fraction of the regression coefficient toward zero. This is
achieved without the need of setting any hyper-parameters.

\begin{figure}
\begin{centering}
\includegraphics[scale=0.1]{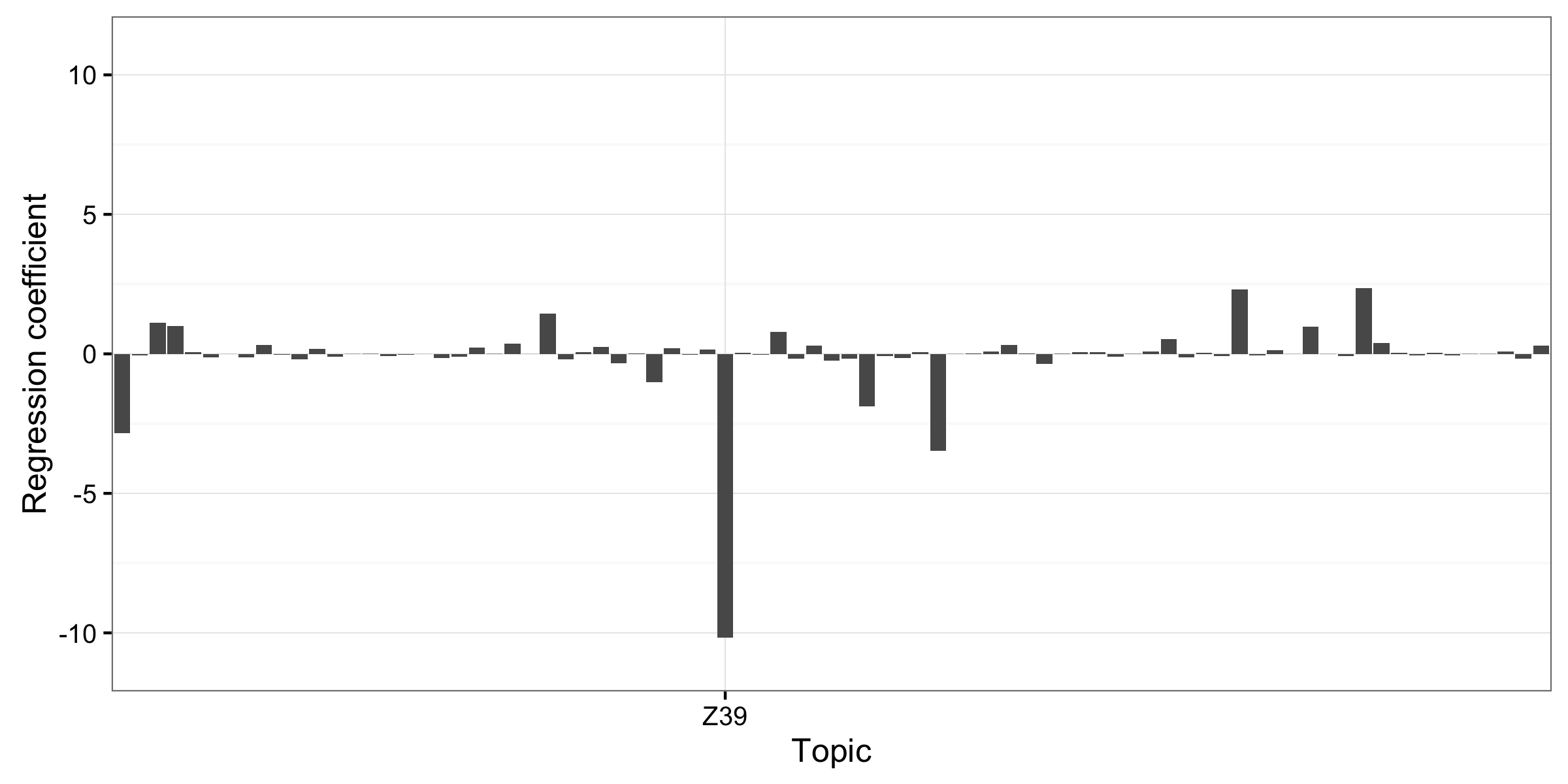}
\par\end{centering}
\begin{centering}
\includegraphics[scale=0.1]{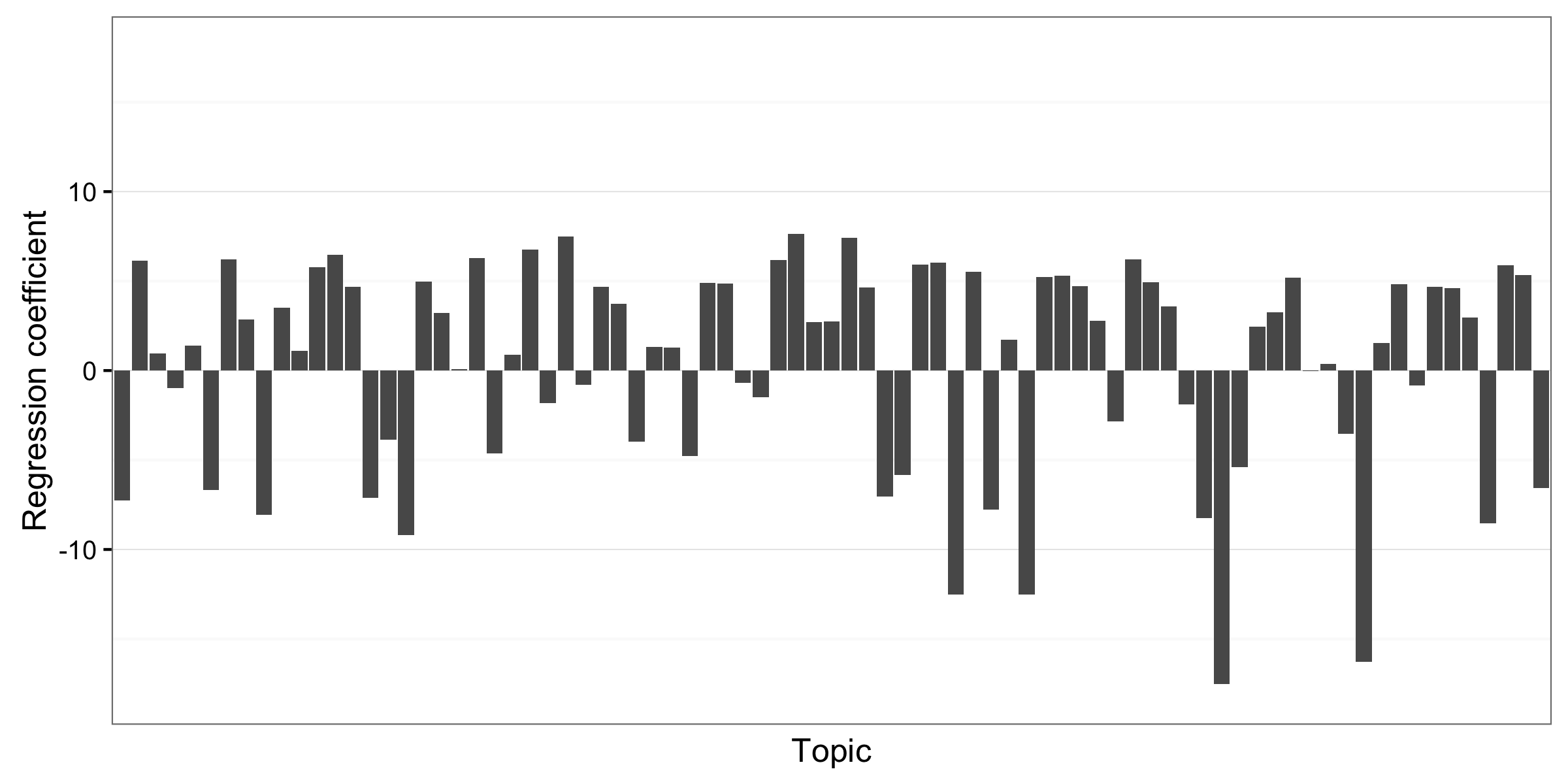}
\par\end{centering}
\caption{Coefficients for the genre \textit{Romance} in the IMDb dataset with
80 topics using the Horseshoe prior (upper) and a normal prior (below).
\label{fig:Hist-beta-1}}
\end{figure}

The Horseshoe shrinkage makes it easy to identify the topics that
affect a given class. This is illustrated for the \textit{Romance}
genre in the IMDb dataset in Figure \ref{fig:Hist-beta-1}. This genre
consists of relatively few observations (only 39 movies), and the
Horseshoe prior therefore shrinks most coefficients to zero, keeping
only one large signal topic that happens to have a negative effect
on the Romance genre. The normal prior on the other hand gives a much
more dense, and therefore much less interpretable solution.

\begin{figure}
\begin{centering}
\includegraphics[scale=0.1]{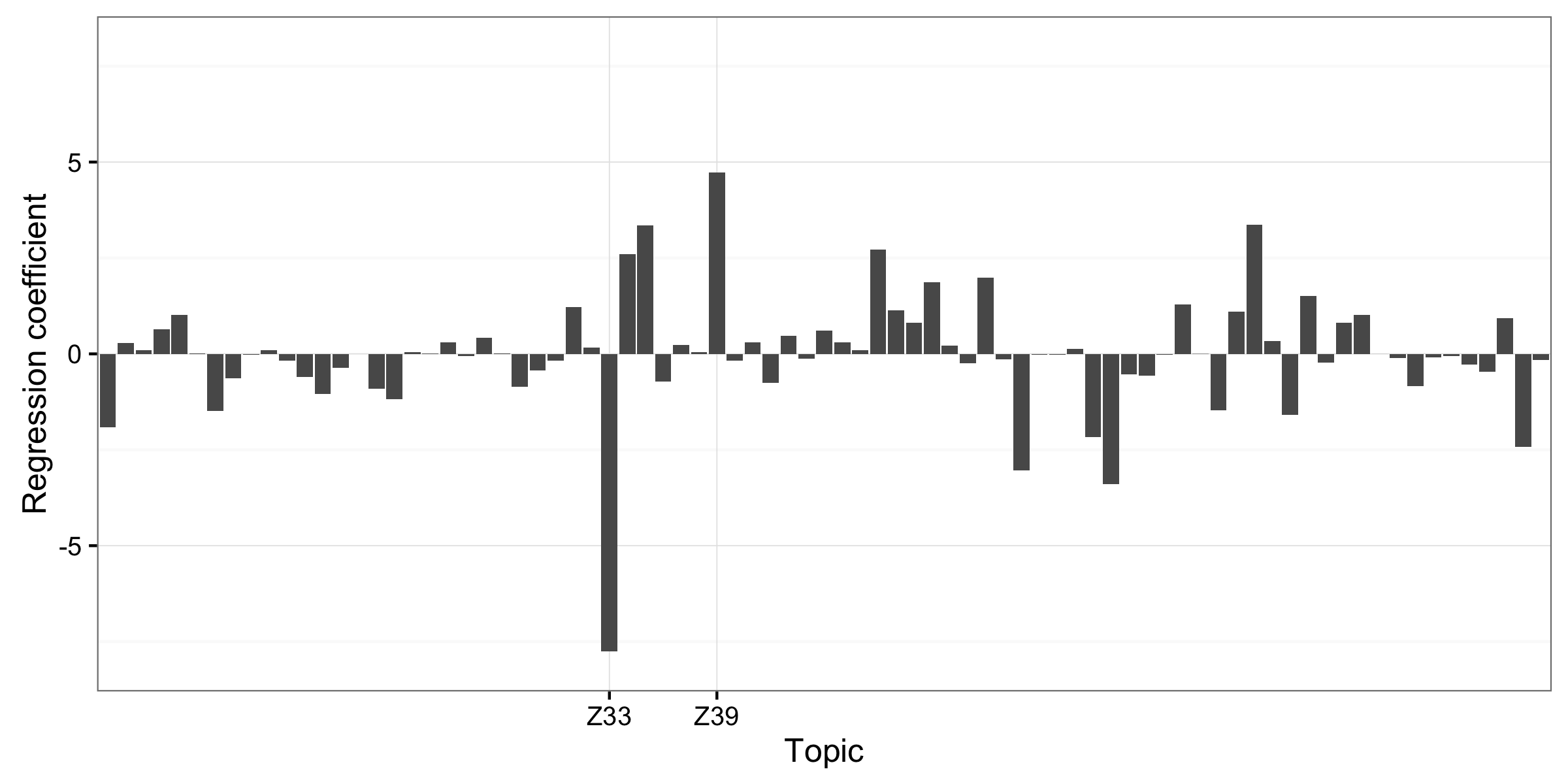}
\par\end{centering}
\begin{centering}
\includegraphics[scale=0.1]{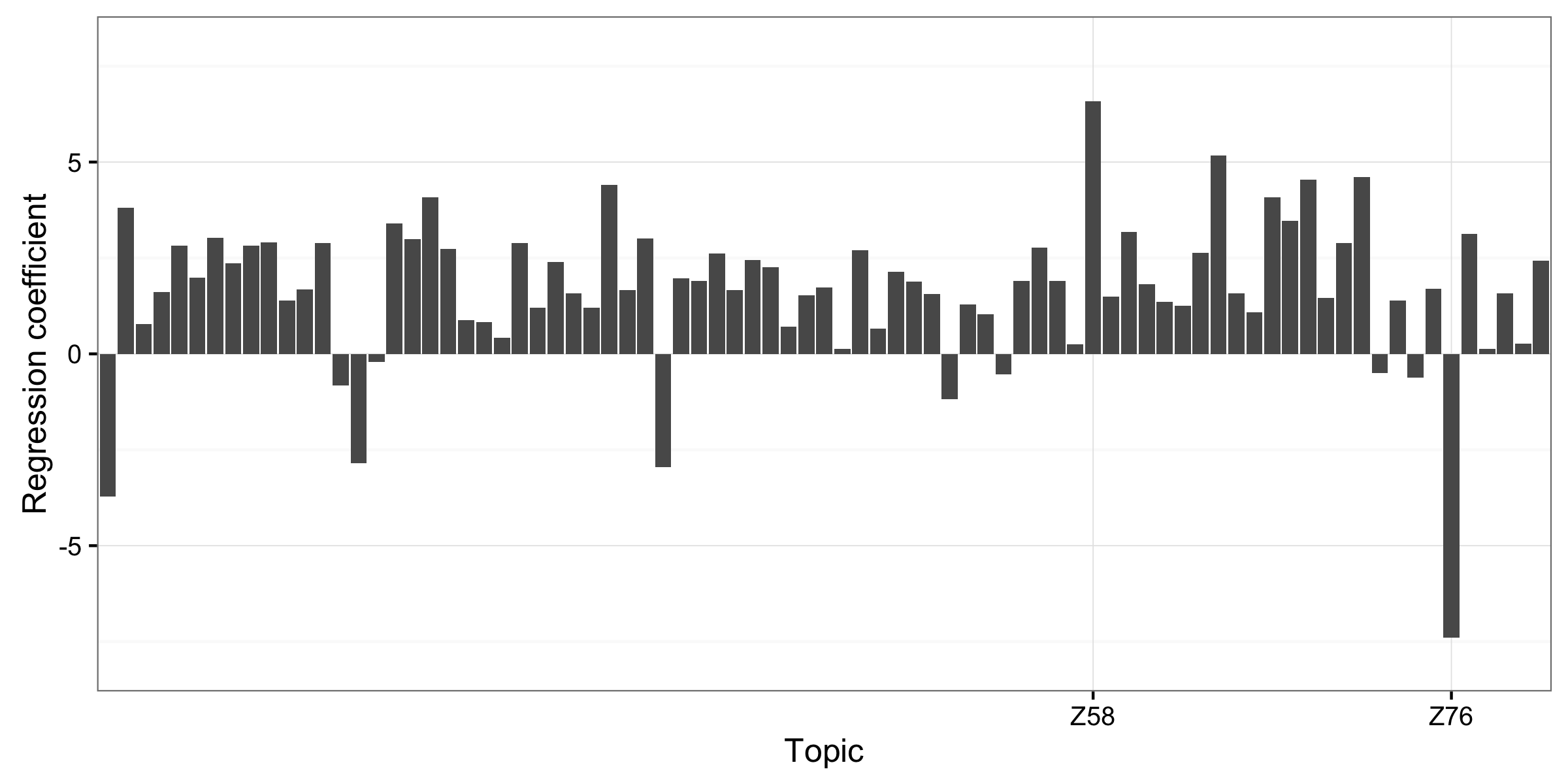}
\par\end{centering}
\caption{Regression coefficients for the class \textit{Crime} for the IMDb
dataset with 80 topics using the Horseshoe prior (upper) and a normal
prior (below). \label{fig:Hist-beta-1-1}}
\end{figure}

Digging deeper in the interpretation of what triggers a Romance genre
prediction, Table \ref{tab:Top-words-in} shows the 10 top word for
Topic 39. From this table it is clear that the signal topic identified
using the Horseshoe prior is some sort of ``crime'' topic that is
negatively correlated with the Romance genre, something that makes
intuitive sense. The crime topic is clearly expected to be positively
related to the \textit{Crime} genre, and Figure \ref{fig:Hist-beta-1-1}
shows that this is indeed the case. 

\begin{table}
\begin{centering}
\begin{tabular}{|c|l|}
\hline 
Topic 33 & earth space planet alien human future years world time mission \tabularnewline
\hline 
Topic 39 & police murder detective killer case investigation crime crimes solve
murdered \tabularnewline
\hline 
\end{tabular}
\par\end{centering}
\caption{Top words in topics using the Horseshoe prior.\label{tab:Top-words-in}}
\end{table}

We can also from Figure \ref{fig:Hist-beta-1-1} see that Topic 33
is negatively correlated with the Crime genre. In Table \ref{tab:Top-words-in}
we can see that Topic 33 seems to be some sort of Sci-Fi topic containing
top words such as ``space'', ``alien'' and ``future''. This
topic has the largest positive correlation with the Sci-Fi movie genre,
which again makes intuitive sense.

\section{Conclusions}

Several supervised topic models have recently been proposed with the
purpose to identify topics that can successfully be used to classify
documents. We have here proposed DOLDA, a supervised topic model with
special emphasis on generating semantically interpretable predictions.
An important component of the model to ease interpretation is the
DO-probit model without a reference class. By coupling the DO-probit
model with an aggressive Horseshoe prior with shrinkage that is allowed
to vary over the different classes it is possible to create a highly
interpretable classification model. At the same time the DOLDA model
comes with very few hyper parameters that needs tuning, something
that is needed in many other supervised topic models such as \citep{jiang2012monte,zhu2012medlda,li2015supervised}.
Our experiments show that the gain in interpretation from using DOLDA
comes with only a small reduction in prediction accuracy compared
to the state-of-the art supervised topic models; moreover, DOLDA outperforms
other fully Bayesian models such as the original supervised LDA model.
It is also clearly shown that learning the topics jointly with the
classification part of the model gives more accurate predictions than
a two step approach where a topic model is first estimated and a classifier
is then trained on the learned topics.

\bibliographystyle{elsarticle-harv}
\addcontentsline{toc}{section}{\refname}\bibliography{articleDObibTex}

\end{document}